\documentclass[conference]{csce}

\usepackage[hmargin=.75in,vmargin=1in]{geometry}
\usepackage[american]{babel}
\usepackage[T1]{fontenc}
\usepackage{times}
\usepackage{caption}

\usepackage{textcomp}
\usepackage{epsfig,graphicx}
\usepackage{xcolor}
\usepackage{amsfonts,amsmath,amssymb}
\usepackage{fixltx2e} % Fixing numbering problem when using figure/table* 
\usepackage{booktabs}

\usepackage{url}
\usepackage{tabu}
\usepackage{algorithm,algorithmic}

\title{
\bf High-Performance Support Vector Machines \\
and Its Applications
}

\author{
    {\bfseries Taiping He, Tao Wang, Ralph Abbey, and Joshua Griffin} \\
    SAS Institute, Inc., Cary, North Carolina, USA 
}

\begin{document}

\maketitle                        %%%% To set Title and Author names.

%%%\begin{abstract}%%%% Replace with your abstract.
%%%Please consider these Instructions as guidelines for preparation of 
%%%Final Camera-Ready papers. The Camera-Ready Papers would be acceptable as 
%%%long as it is formatted reasonably close to the format being suggested here. 
%%%Note that these instructions are reasonably comparable to the standard IEEE 
%%%typesetting format. Type the abstract (100 words minimum and 150 words maximum) 
%%%using Italic font with point size 10. The abstract is an essential part of the 
%%%paper. Use short, direct, and complete sentences. It should be brief and as concise as possible.
%%%\end{abstract}

\begin{abstract}
The support vector machines~(SVM) algorithm is a popular classification technique in data mining and machine learning. 
In this paper, we propose a distributed SVM algorithm and demonstrate its use in a number of applications. 
The algorithm is named high-performance support vector machines~(HPSVM).
The major contribution of HPSVM is two-fold. 
First, HPSVM provides a new way to distribute computations to the machines in the cloud without shuffling the data.
Second, HPSVM minimizes the inter-machine communications in order to maximize the performance.
We apply HPSVM to some real-world classification problems and compare it with the state-of-the-art SVM technique implemented in R on several public data sets.  HPSVM achieves similar or better results. 
\end{abstract}

\vspace{1em}
\noindent\textbf{Keywords:}
{\small  Support vector machines; 
Distributed data mining; 
Classification; 
Big data algorithms}

%%% {\small  A maximum of 6 keywords} %%%% Replace with your keywords

%%%%%%%%%%%%%%%%%%%%%%%%%%%%%%%%%%%%%%%%%%%%%%%%%%%%%%%%%%%%

\section{Introduction}

Support vector machines~(SVM) \cite{cortes1995, vapnik1995} 
are powerful machine-learning techniques for classification. Since 1995, experts in the machine-learning community have shown significant interest in SVM.

Like many predictive modeling algorithms, SVM consists of training, validating, and testing stages. The training stage involves solving a dense quadratic programming or dense convex optimization problem. 
Since the size of the quadratic problem depends on the total number of observations, 
general-purpose quadratic programming solvers are not competitive without specialization, 
especially when the training data are relatively large.

In order to solve large data set problems, several algorithms have been proposed in the literature. 
Among them are active-set algorithms \cite{joachims1998, osuna1997} 
and sequential minimal optimization~(SMO) \cite{platt1999}. 
The idea of the active-set algorithms is to decompose the big problem into a series of small tasks. 
The decomposition splits the training data set into inactive and active parts. The active part is called the ``working set'' and is normally small. The solver focuses on the working set and keeps the support vectors in the subsequent working set.
In fact, SMO is a special case of the active-set algorithm, where the size of the working set is $2$. 
Active-set algorithms have their own limitations.
Because the training for active-set algorithms is sequential and the next iteration depends on the previous results, the problem cannot be easily separated for parallel processing. 
Moreover, memory and CPU usage increase rapidly as the number of support vectors grows during the training. 

In 2003, Ferris and Munson \cite{ferris2003} proposed the interior-point method by applying the Sherman-Morrison-Woodbury \cite{wright2000} formula, which makes it possible to solve very large problems.
In \cite{ferris2003},  the computation of the large matrix is transformed to the computation of a small core matrix through the linear algebra technique. With this technique, the memory required for the quadratic problem is reduced from $O(n^2)$ 
to $O(n) + O(m^2)$, where $n$ is the number of observations and $m$ is the number of features.

Since 2008, several parallel or distributed computation algorithms have been proposed in the literature. 
Gertz and Griffin \cite{griffin2008} proposed an interior-point algorithm and implemented it with the object-oriented package OOQP \cite{ooqp}. 
%The algorithm was further improved in \cite{griffin2008}.
Chang et al. \cite{chang2008} applied the interior-point method in a distributed computing environment. 
Chang et al. handled the kernel matrix by a low-rank approximation that uses partial Cholesky decomposition with pivoting, and  major computation of the SVM training is performed in the distributed processors. 
Woodsend and Gondzio \cite{woodsend2009} 
proposed a Hybrid MPI/OpenMP parallel algorithm,
%\cite{hybrid}, 
which uses the interior-point algorithm, avoids the dense Hessian matrix, and computes a distributed Cholesky decomposition. The authors claimed that their approach was much better than others during the 
PASCAL Challenge competition \cite{pascal}. 
%%%
Unfortunately, one important issue has not been fully discussed in the literature: 
the use of distributed vectors and distributed vector algebra. Another issue is that the inter-machine communication
makes the implementation of distributed SVM difficult.

In this paper, we propose a distributed algorithm to solve the large primal-dual SVM problem. 
The distributed SVM algorithm is called high-performance support vector machines~(HPSVM).
%We focus on the linear SVM with distributed computation. Several reasons lead this algorithm very important and practically useful in the field. 
We consider a few important issues in the design of this algorithm.
The first concern is the model training time and memory usage. For a large data set, we need an algorithm that can train the model in a reasonable amount of time and use a limited amount of run-time memory. 
The second concern is the model storage and easy scoring process. The model itself should not be too big to store even if the number of support vectors is large. And the scoring process should be simple. Therefore, finding an algorithm which can train a good model on a large data set
%billions of observations 
in a reasonable amount of time and provide an easy scoring mechanism is critical in the SVM application field. At the same time, it is essential that a good algorithm should be able to take the advantage of cloud computing and the distributed Hadoop file system. 

In our implementation, we adopted message passing interface~(MPI) for the communication between the master node and the worker nodes.
We had two principles in mind when we designed the HPSVM algorithm.
First, we knew that data shuffling in the distributed environment can be very expensive.
Secondly, we knew that inter-machine communication could also significantly slow down the entire process.
This paper offers two major contributions.
First, we propose a new way to distribute computations to the machines in the cloud.
Second, we minimize the communications among the machines in the cloud to maximize the performance.
We carefully designed the algorithm so that data shuffling is not required and inter-machine communications are minimized.
In other words, all data that are saved in each worker node are loaded locally.
%We also include a complexity analysis of the distributed SVM algorithm.
%Experiments demonstrate that HPSVM scales very well, and can handle large data sets with good results. 
%%%   ??????????????

The rest of the paper is organized as follows. Section 2 briefly introduces support vector machines. Section 3 talks about the interior-point method. Section 4 presents the distributed SVM algorithm. Section 5 provides a complexity analysis of the algorithm. 
Experiments and their results are shown in Section 6. We draw conclusions in Section 7.

\section{Support Vector Machines}

%Let's go over the basic SVM theory.
%Provde the primal and dual objective functions and basic theory.
%Provde the detailed notation for each letters.

%*******************************

In this section, we provide the basic notations used in this paper and describe the SVM classification concept and some formulas. 

First we describe the basic notation.
Let $m$, $n$ be positive integers. In this paper, $n$ is the number of observations and $m$ is the number of features. We assume that $m < n$. For each $i=1,\ldots,n$, $x_i^T\in R^m$. 
We have $X^T=(x_1^T, x_2^T,\cdots,x_n^T)$, and $X\in R^{n\times m}$. 
Let $d=(d_1,\cdots,d_n)^T\in R^n$, $d_i\in \{-1, 1\}$ for each $i=1, \ldots,n$. 
Let $w\in R^m$, $\beta\in R$, $e=(1,\dots, 1)^T\in R^n$, and $s, u, v, z\in R^n$. Let $D$ denote the $n\times n$ diagonal matrix $diag(d)$, and similarly denote the $n\times n$ matrices 
$S=diag(s)$, $U=diag(u)$, $V=diag(v)$, 
and 
$Z=diag(z)$. 
The training data set is $(X, d) \subset R^{n\times (m+1)}$. Each row of the training data $(x_i, d_i)$ represents a single observation.
The size of the matrices $S$, $U$, $V$, and $Z$ is $n\times n$. 

A support vector machines~(SVM) algorithm is a classification algorithm that provides the mapping between the feature space and the target labels. The hyperplane $w^Tx - \beta = 0$ is used to define the mapping. 
%The decision function here $f(x) = w^Tx - \beta$ where $x$ is a point in the feature space, $w$ is the parameter and $\beta$ is the interception.
The training of the SVM is to find $w$ and $\beta$ such that the maximum margin $2/||w||$ between the two hyperplanes $w^Tx-\beta=-1$ and $w^Tx-\beta=1$ is reached.
The decision function 
\begin{equation}
   f(x) = sign(w^Tx - \beta)
\end{equation}
defines the classifier, where the values $1$  and $-1$ are mapped to the target labels.

%\vskip 0.5in
The optimization problem is to find $w$, $\beta$, and $z$ that satisfy
\begin{equation} \label{primal}
\begin{array}{lll}
    \text{minimize}  & \frac{1}{2} w^Tw + \tau e^Tz  \\
    \text{subject to} & D(X^Tw - \beta e) \ge e - z, \\
                   & z \ge 0
\end{array}
\end{equation}
where $\tau > 0$ is a penalty parameter and $z$ is a slack variable. We call equation (\ref{primal}) the primal problem.

%\vskip 0.5in
%The Lagrangian function associated with (\ref{primal}) is
%\begin{equation}
%    L_P(w, \beta, z, v) = \frac{1}{2}wT^w + e^Tz + v...
%\end{equation}
%where ??? are Lagrangian multipliers.

%\vskip 0.5in
The dual problem of the primal problem (\ref{primal}) is
\begin{equation} \label{dual}
\begin{array}{lll}
    \text{minimize}  & \frac{1}{2} v^TDX^TXDv - e^Tv \\
    \text{subject to} & e^TDv = 0, \\
                   & 0\le v \le \tau e
\end{array}
\end{equation}

%\vskip 1in
%The Lagrangian function associated with \ref{equ2} is:
%\begin{equation}
%    L_D(v, s, u) = \frac{1}{2}v^TDX^TXDv - e^Tv + S*.... 
%\end{equation}
%where ??? are Lagrangian multipliers.

%\vskip 0.5in
If we replace the $n\times n$ matrix $DX^TXD$ with $Q$, the generalized problem becomes
\begin{equation} \label{dualq}
\begin{array}{lll}
    \text{minimize}  & \frac{1}{2} v^TQv - e^Tv \\
    \text{subject to} & e^TDv = 0, \\
                   & 0\le v \le \tau e
\end{array}
\end{equation}

A nonlinear kernel can also be introduced when we solve the quadratic program (\ref{dualq}).  
Let $K(\cdot,\cdot)$ be a function $K$: $R^n\times R^n \to R$, 
$q_{ij}=d_iK(x_i,x_j)d_j$ for $i,j=1,\ldots,n$. 
Then $Q=(q_{ij})_{n\times n}$ is a dense $n\times n$ matrix. We call $K$ the kernel function.
%becomes a dense positive defined matrix. Thus the optimization problem (\ref{dualq}) is well-defined.
The frequently used kernel functions include the polynomial function, the radial basis function~(RBF),  the sigmoid function, and so on. 
If the kernel function satisfies Mercer's condition \cite{merc1909}, then the resulting kernel matrix $Q$ is symmetric positive-definite. Thus the quadratic problem (QP) is convex and the global solution exists.

\section{Interior-Point Method}

Many research publications focus on solving the dual problem or the nonlinear kernel mapping problem (\ref{dualq}). 
The reason is that the optimization process tries to solve a simple quadratic problem with basic linear constraints, which is easier than solving the primal problem.
Implementation issues arise as the training data become large. 
Since the matrix $Q$ size is $n\times n$ and dense, the memory requirement $O(n^2)$ makes it very difficult for a regular solver to handle.
To resolve this problem, Ferris and Munson \cite{ferris2003} apply the Sherman-Morrison-Woodbury \cite{wright2000} formula  and transform a large $n\times n$ matrix to a small core matrix of size $m\times m$. 
With this technique, the memory required for the solver is reduced from $O(n^2)$ to $O(nm)$, or reduced even further to $O(n) + O(m^2)$ if the data are not loaded into memory.

%%%%%%%% The SWM formula specifies that  ... , the principle is used here.
%% Need to add the SMW formula and discuss the relation of the interior-point method and the formula
%%  This part can be done at the end of the section.

%%%% Need to mention that only the linear kernal is discussed in the algorithm.

%%%% For low degree polynomial cases, the mapping of the dimentions are performed first before the training.

The development of the interior-point method involves the primal-dual Lagrangian function, which is associated with equations (\ref{primal}) and (\ref{dual}),
\begin{align}
    L(w, \beta, v, s, u, z) = \frac{1}{2}w^Tw+\tau e^Tz - v^T(D(Xw-\beta e) \nonumber \\
          -e+z) -u^Tz -s^Tv
\end{align}
where $s$, $u$, and $v$ are Lagrangian multipliers.
The primal-dual problem is to solve the following system. 
(For details, see Nocedal and Wright \cite{wright2000}, chapter 19, and \cite{ferris2003}.)
%here we adopt the notation from \cite{griffin2008}. 
\begin{align}
    w - Y^Tv = r_w = 0 \label{lag1}\\
    d^Tv = \rho_\beta = 0 \\
    \tau e - v -u = r_z = 0 \\
    Yw-\beta d - e + z - s = r_v = 0 \\
%    ZUe = r_u = 0 \\
    ZUe = r_u = 0 \label{lag2zu}\\
    SVe = r_s = 0  \label{lag2}\\
    s,\ u,\ v,\ z \ge 0  \nonumber 
\end{align}
where $Y=DX$ is an $n\times m$ matrix and $S$, $U$, $V$, and $Z$ are diagonal matrices as defined in the previous section.
The interior-point algorithm approaches the perturbed equation (\ref{lag2zu}) and equation (\ref{lag2}) so that the variables
$(s,u,v,z)$ remain positive and approach zero only in the limit. For brevity, we perform the block reduction steps
on the unperturbed equations, but the steps in the perturbed case are analogous.
The Newton system of equations (\ref{lag1}) to (\ref{lag2}) is
\begin{align}
    \Delta w - Y^T\Delta v = - r_w                         \label{newton1} \\
    d^T\Delta v = -\rho_\beta                              \label{newton2} \\
    -\Delta v - \Delta u = -r_z                            \label{newton3} \\
    Y\Delta w - d\Delta \beta + \Delta z -\Delta s = -r_v  \label{newton4} \\
    Z\Delta u + U\Delta z = -r_u                           \label{newton5} \\
    S\Delta v + V\Delta s = -r_s                           \label{newton6}
\end{align}

From the Newton system, combining equations (\ref{newton3}) and (\ref{newton5}) to eliminate $\Delta u$, we obtain
\begin{equation} \label{equation18}
    -\Delta v + Z^{-1}U\Delta z = - \hat{r_z}
\end{equation}
where $\hat{r}_z = r_z + Z^{-1}r_u$. Combining equations (\ref{newton4}) and (\ref{newton6}) to eliminate $\Delta s$, we obtain
\begin{equation} \label{equation19}
Y\Delta w - d\Delta\beta + \Delta z + V^{-1}S\Delta v = -\hat{r}_v 
\end{equation}
where $\hat{r}_v = r_v + V^{-1}r_s$. Furthermore, we remove $\Delta z$ from equations (\ref{equation18}) and (\ref{equation19}):
\begin{equation} \label{eq22}
    Y\Delta w - d\Delta\beta + (U^{-1}Z + V^{-1}S)\Delta v = -\hat{r}_v + U^{-1}Z\hat{r}_z
\end{equation}
We denote $\Omega = U^{-1}Z + V^{-1}S$ and $r_\Omega=\hat{r}_v-U^{-1}Z\hat{r}_z$. Thus equation (\ref{eq22}) becomes
\begin{equation} \label{eq23}
    Y\Delta w - d\Delta\beta + \Omega\Delta v = -r_\Omega
\end{equation}
Combining equation (\ref{eq23}) with equations (\ref{newton2}) and (\ref{newton1}), we have
\begin{eqnarray}
    \left\{
        \begin{array}{rrrrrll}
            \Omega \Delta v & - & d\Delta\beta & + &Y\Delta w & = & -r_\Omega    \\
            d^T\Delta v & & & & & = & -\rho\beta   \\
            -Y^T\Delta v & & & + &\Delta w & = & -r_w
        \end{array}
    \right.
\end{eqnarray}
Applying simple Gaussian elimination, we have
\begin{eqnarray} \label{eq24}
    \left\{
        \begin{array}{rrll}
%            \Omega \Delta v & - & d\Delta\beta & + &Y\Delta w & = & -r_\Omega    \\
%                            &  & d^T\Omega^{-1}d\Delta\beta &- &d^T\Omega^{-1}Y\Delta w  & = & -\hat{\rho}_\beta   \\
%                            & - & Y^T\Omega^{-1}d\Delta\beta &+&(Y^T\Omega^{-1}Y+I)\Delta w & = & -\hat{r}_w 
            \Omega \Delta v  -  d\Delta\beta  + &Y\Delta w & = & -r_\Omega    \\
                               d^T\Omega^{-1}d\Delta\beta - &d^T\Omega^{-1}Y\Delta w  & = & -\hat{\rho}_\beta   \\
                             -  Y^T\Omega^{-1}d\Delta\beta +&(Y^T\Omega^{-1}Y+I)\Delta w & = & -\hat{r}_w 
        \end{array}
    \right.
\end{eqnarray}
where $\hat{\rho}_\beta = \rho_\beta - d^T\Omega^{-1}r_\Omega $ and $\hat{r}_w=r_w+Y^T\Omega^{-1}r_\Omega$. Eliminating $\Delta\beta$ from the last equation of (\ref{eq24}), we obtain
\begin{eqnarray} \label{equations}
    \left\{
        \begin{array}{rrrrrll}
            \Omega \Delta v & - & d\Delta\beta & + &Y\Delta w & = & -r_\Omega    \\
                            &  & \sigma\Delta\beta &- &d^T\Omega^{-1}Y\Delta w  & = & -\hat{\rho}_\beta   \\
                            &   &  & &M\Delta w & = & -\hat{r}_M 
        \end{array}
    \right.
\end{eqnarray}
%where $\sigma = d^T\Omega^{-1}d$ and
where 
\begin{align}
    M = I + Y^T\Omega^{-1}Y -\frac{1}{\sigma}Y^T\Omega^{-1}dd^T\Omega^{-1}Y \label{MM} \\
    \hat{r}_M = \hat{r}_w - \frac{1}{\sigma}Y^T\Omega^{-1}d\hat{\rho}_\beta \label{rM}
\end{align}
and $\sigma = d^T\Omega^{-1}d$.

After obtaining $\Delta w$ from (\ref{equations}), we substitute it back into the first two equations, and we have
\begin{align}
    \Delta \beta = \frac{1}{\sigma}(-\hat{\rho}_\beta + d^T\Omega^{-1}Y\Delta w) \label{deltabeta} \\
    \Delta v = \Omega^{-1}(-r_\Omega + d\Delta \beta - Y\Delta w) \label{deltav}
\end{align}
Thus we complete one iteration of the Newton system.

%\vskip 1in

\section{Distributed SVM Algorithm}

%We present the distributed SVM algorithm in this section.

Assuming that the data are separated into $p$ blocks and distributed to $p$ worker nodes, we have 
%$X = (X_1^T, X_2^T,\cdots,X_p^T)^T$. Here 
\begin{align}
    X  =  \begin{pmatrix}
        X_1 \\ X_2 \\ \vdots \\ X_p
    \end{pmatrix}
    & \text{and} \ Y  =  DX = \begin{pmatrix}
        Y_1 \\ Y_2 \\ \vdots \\ Y_p
    \end{pmatrix}
\end{align}
The matrix $\Omega$ is also diagonal and can be written in the form
\begin{align}
    \Omega  =  diag(\Omega_1,\Omega_2,\cdots,\Omega_p)  = 
    \begin{pmatrix}
        \Omega_1 &         &       & \\
                 &\Omega_2 &       & \\
                 &         & \ddots      & \\
                 &         &       & \Omega_p
    \end{pmatrix}
\end{align}

For the second term of matrix $M$ in (\ref{MM}), we have
\begin{align}
    Y^T\Omega^{-1}Y &= (Y_1^T Y_2^T \cdots Y_p^T)
    \begin{pmatrix}
        \Omega_1^{-1} &         &       & \\
                      &\Omega_2^{-1} &       & \\
                 &         & \ddots      & \\
                 &         &       & \Omega_p^{-1}
    \end{pmatrix}
    \begin{pmatrix}
                 Y_1 \\ Y_2 \\ \vdots \\ Y_p
    \end{pmatrix} \nonumber \\
    & = Y_1^T\Omega_1^{-1}Y_1 + Y_2^T\Omega_2^{-1}Y_2 + \cdots + Y_p^T\Omega_p^{-1}Y_p \nonumber \\
    & = \sum_{i=1}^{p} Y_i^T\Omega_i^{-1}Y_i \label{eqYWY}
%    & = \sum_{i=1}^{p} Y_i^T\Omega_i^{-1}Y_i
\end{align}
Similarly, for the third term of matrix $M$, we have
\begin{equation}
    Y^T\Omega^{-1}dd^T\Omega^{-1}Y = (\sum_{i=1}^{p}Y_i^T\Omega_i^{-1}d_i)(\sum_{i=1}^{p}d_i^T\Omega_i^{-1}Y_i)
\end{equation}
where
\begin{align}
    d=\begin{pmatrix}
        d_1 \\ d_2 \\ \vdots \\ d_p
    \end{pmatrix}
\end{align}
The calculation of $\sigma$ is the same:
\begin{equation}
    \sigma = d^T\Omega^{-1}d = \sum_{i=1}^p d_i^T\Omega_i^{-1}d_i
\end{equation}

We see that for each worker node $i$, the data $X_i$ or $Y_i$ stays in that local node and never moves to other nodes.
Each worker node $i$ performs its computation with the corresponding parts $Y^T_i\Omega_i^{-1}Y_i$, $Y_i^T\Omega_i^{-1}d_i$, and $d_i^T\Omega_i^{-1}d_i$, and then the results are gathered to the master node through the all-reduce actions of the MPI interface. The communication traffic size from the worker node to the master node is $O(m^2) + O(n/p)$. We see that only the master node holds the matrix $M$ and the residual $\hat{r}_M$ of equation (\ref{rM}). 

Once the $M$ and $\hat{r}_M$ are ready, the master node computes the $\Delta w$ from the third equation in (\ref{equations})
and then computes $\Delta \beta$ from equation (\ref{deltabeta}), after which the master node broadcasts $\Delta w$ and $\Delta\beta$ to each worker node. The network traffic for this action is only $O(m+1)$.

Let us look at $\Delta v$. From equation (\ref{deltav}), we have 
\begin{align}
    \Delta v & = 
    \begin{pmatrix}
        \Delta v_1 \\ \Delta v_2 \\ \vdots \\ \Delta v_p
    \end{pmatrix}
%    \\
     = \begin{pmatrix}
                \Omega_1^{-1} &    &    &       \\
                  &\Omega_2^{-1}        &    &   \\
                  &    & \ddots       &   \\
                  &    &    &    \Omega_p^{-1}   \\
        \end{pmatrix} \\
        & 
        \begin{pmatrix}
          - \begin{pmatrix}
            r_{\Omega_1} \\ r_{\Omega_2} \\ \vdots \\ r_{\Omega_p}
            \end{pmatrix}
        + \begin{pmatrix}
            d_1 \\ d_2 \\ \vdots \\ d_p
         \end{pmatrix}
         \Delta \beta
        -\begin{pmatrix}
            Y_1 \\ Y_2 \\ \vdots \\ Y_p
         \end{pmatrix}
         \Delta w
    \end{pmatrix}
\end{align}
Then each worker node calculates its own portion of $\Delta v$,
\begin{equation}
    \Delta v_i = \Omega_i^{-1} (-r_{\Omega_i} + d_i\Delta\beta-Y_i\Delta w)
\end{equation}
and calculates $\Delta u_i$, $\Delta s_i$, and $\Delta z_i$. After that, it updates $\Omega_i$ and corresponding residues.

The iteration finishes when $\Delta w$ and $\Delta \beta$ are small and meet the stopping criteria. The support vectors are held in each worker node, and the information is reported to the master node. Therefore, the master node holds all model information and parameters.

%\vskip 2in

%In this section, the detailed algorithm steps for Master and Work nodes are provided.
 %for some information please see: \cite{1}, also \cite{1,2,3}.

Now let us summarize the algorithm for the distributed Newton method. Here we have the computing system, which consists of one master node and multiple worker nodes. The communication between the master node and the worker nodes occurs through the modified MPI.
The distributed SVM algorithm is specified in Algorithm \ref{alg:svm}.
\begin{algorithm}
  \caption{Distributed SVM algorithm }
  \label{alg:svm}
  \begin{algorithmic}[1]
          \STATE \label{item1} Distribute data to worker nodes from the master node.
      \STATE \label{item2} Start the Newton iteration process on both the master node and the worker nodes.
          \FOR{\label{item3} each worker node $i$}
          \STATE Compute           
          $Y_i^T\Omega_i^{-1}Y_i$, $Y_i^T\Omega_i^{-1}d_i$, $r_{\Omega_i}$,
          $d_i^T\Omega_i^{-1}r_{\Omega_i}$, and $d_i^T\Omega_i^{-1}d_i$.
          \ENDFOR
          \STATE \label{item4} Do the all\_reduce actions to obtain the results on the master node
         for $Y^T\Omega^{-1}Y$ and $Y^T\Omega^{-1}d$. 
         \STATE On the master node, calculate the $\hat{\rho}_\beta$ and $\hat{r}_w$.
      \STATE \label{item5} Solve the third equation of (\ref{equations}) on the master node to get $\Delta w$,  and then compute 
        $\Delta\beta$ from equation (\ref{deltabeta}).
      \STATE \label{item6} The master node checks the stopping criteria. Go to step \ref{itemend} if the stopping tolerance is reached.
          \STATE \label{item7} The master node broadcasts $\Delta w$, $\Delta \beta$, and $\sigma$ to each worker node.
          \FOR {\label{item8} each worker node $i$}
      \STATE  Calculate $\Delta v_i$, $\Delta u_i$, $\Delta s_i$, $\Delta z_i$, 
        and the corresponding residues.
        \ENDFOR
    \STATE Go to step \ref{item3}.
    \STATE Return and output results. \label{itemend}
  \end{algorithmic}
\end{algorithm}
It is worth mentioning that our algorithm also applies in single-machine mode. 
Actually the approach of \cite{griffin2008} is a special case of our algorithm. 
In this case, both the master node and the worker node exist in the same machine.
You can easily see that the bottleneck is clearly the calculation in equation (\ref{eqYWY}): 
for big data applications, even vector addition becomes prohibitive if storage and calculation occurs on a single node.   
For example, suppose one data set has a billion observations.  Merely computing the vectors $(s,u,v,z)$ and work vectors  $(\Delta s, \Delta u,\Delta v, \Delta z)$ on a single machine would require 64G of RAM. 
Thus, even if the equation (\ref{eqYWY}) is parallelized, 
the mere calculations of equations (\ref{newton1})--(\ref{newton2}), if performed on a single node in serial,
would bottleneck the approach for big data problems.  In short, any entity whose size is on the order of the number
of observations must be stored and updated in distributed fashion. With recent cloud computing technology and the distributed Hadoop file system, the importance of our distributed algorithm is obvious.

\section{Complexity Analysis of the Algorithm}
We now take a look of the detailed complexity of the algorithm, including memory usage and CPU time.

\subsection{Memory Usage}

%Now let us investigate the memory usage of the training process.

Here we have $n$, the number of total observations; $m$, the number of features;  and $p$, the number of worker nodes used.
Assume the data are evenly distributed among the worker nodes. 
From step \ref{item1} of the Algorithm \ref{alg:svm} in the previous section, we see that the data size in a worker node is $O(nm/p)$. Assume that all the data are loaded into the memory during the training. From step \ref{item3}, the memory that is needed to hold matrices and residues is $O(m^2) + O(n/p)$. The memory required for step \ref{item4} is $O(m^2)$. Thus the total memory size for the training in each worker node is
\begin{equation}
    O(\frac{mn}{p}) + O(m^2) + O(\frac{n}{p}) = O(\frac{mn}{p}) + O(m^2)
\end{equation}
When $m < n/p$, the total memory used for each worker node is $O(nm/p)$.
On the other hand, from step \ref{item4}, the memory needed for the master node is $O(m^2) + O(n/p)$.

For example, if the total number of observations $n=1$ billion, the number of features $m=1,000$, and the number of worker nodes 
$p=1,000$, then for each worker node the memory that is needed for the training in each worker node is
\begin{equation}
    \frac{m* n}{p}* 8\ \text{byte} = \frac{1000*10^9}{1000}*8\ \text{byte} = 8*10^9\ \text{byte} = 8 \text{GB}
\end{equation}

%In the actual implementation, we do not have to load all the data into memory. For step \ref{item1}, the data can be saved to a local file system and loaded online at step \ref{item3} when the matrix $Y_i^T\Omega_i^{-1}Y_i$ is calculated. 
%Thus the memory that is needed for each worker node is only $O(m^2) + O(n/p)$. Actually, when we implement the algorithm, we also need to consider the overall time usage; it is a trade-off for the memory. When the data are saved in the local file system, the I/O time can be large, especially when there are multiple iterations in the Newton process and each iteration accesses the whole data set in the local file system.

In our implementation, the whole data set is loaded into memory to improve speed and performance. 
%It has been our experience that, in a parallel environment, the cumulative RAM allocated across dedicated nodes can usually be selected to exceed the actual problem sizes that we encounter.  
Note that this paradigm could easily be revised to read data in pages of memory if necessary.

%Each work node memory needed.

%The master node memory needed.

%Other buffer memory needed for communication.

\subsection{CPU Time Analysis}

%We now look at the CPU time usage during the training.

For step \ref{item3}, the time to compute $Y_i^T\Omega_i^{-1}Y_i$ is $O(nm^2/p)$. For step \ref{item4}, the time to perform the all\_reduce step is $O(m^2p)$. For step \ref{item5}, the time to solve the equation $M\Delta w = \hat{r}_M$ is $O(m^3)$. And for step \ref{item8}, the time needed is $O(n/p)$. Therefore the total time needed for each Newton iteration is
\begin{equation} \label{equation35}
    O(\frac{nm^2}{p}) + O(m^3) + O(\frac{n}{p}) = O(\frac{nm^2}{p}) + O(m^3)
\end{equation}
When $n/p > m$, from equation (\ref{equation35}) we see that for each Newton iteration, 
the total CPU time is $O(nm^2/p)$.

Actually, for step \ref{item5} of Algorithm \ref{alg:svm}, we can apply different techniques to solve the equation $M\Delta w=\hat{r}_M$. 
The time needed to solve the equation can be reduced from $O(m^3)$ to $O(C_0m^2)$ for some constant value $C_0$.

Suppose the number of Newton iterations is $k_0$. Then the total CPU time needed is $O(k_0nm^2/p)$. 
In the actual implementation, the multiple-thread programming technique can be applied. In this case, if there are $q$ processors in 
each worker node, the total time can be further reduced to $O(k_0nm^2/pq)$.

%\subsection{Universal Data Feeder~(UDF) for Data Integration and Data Access}
\subsection{Data Integration and Data Access}

The training data can be saved on a local disk; in a distributed Hadoop file system; in a distributed database system 
such as Teradata \cite{teradata}, Greenplum \cite{greenplum}, or Aster \cite{aster}; and so on.
For commercial distributed database systems~(such as Teradata, Greenplum, and Aster), the data can be saved on the same worker nodes and can be loaded locally on the fly during the training. 
This is very important for distributed algorithms: each worker node first computes on its own data, and data movement between work nodes should not happen unless it is necessary. 
Therefore the data access time can be reduced and the network communication time can also be reduced.
In fact, this is one of the most commonly used data access methods in commercial environments.
Our HPSVM implementation is currently running on a wide range of platforms including Hadoop, 
Teradata, Greenplum, Aster, and many others.
In this section, we will discuss the data integration and data access strategy that allows HPSVM to run on those platforms successfully.

Our distributed SVM algorithm can run in two modes: 
symmetric multiple processing~(SMP) mode and massively parallel processing~(MPP) mode. 
The following paragraphs briefly introduce these two computing modes.

In SMP mode, multiple CPUs~(cores) are controlled by a single operating system, and the resources~(such as disks and memory) are shared in the machine. Our algorithm uses multiple concurrent threads in SMP mode in order to take advantage of parallel execution. In SMP mode, you have the flexibility to choose to run our algorithm with a single thread or multiple threads. By default, SMP uses the number of CPUs~(cores) on the machine to determine the number of concurrent threads. You can also specify the number of threads to overwrite the default. 

In MPP mode, multiple machines in a distributed computing environment~(cloud) participate in the calculations. Because we chose to use MPI, the assumption is that the resources~(such as disks and memory) are shared only within each machine, not between the machines. One machine communicates to another machine through MPI. 
In MPP mode, you can run a single thread or multiple threads on a single machine or multiple machines. By default, all the available machines in the distributed computing environment are used, and the number of CPUs~(cores) on each machine determines the number of concurrent threads. You can also specify the number of machines or threads to overwrite the default.

We deploy a comprehensive data integration and data access strategy to support the two computing modes. In this strategy, a universal data feeder~(UDF) is used between HPSVM and the platform. Our UDF supports a variety of platforms including Hadoop, 
Teradata, Greenplum, Aster, and many others.
This UDF has two data access methods: the SMP data access method and the MPP data access method.

In SMP mode, the UDF supports the SMP data access method. 
The data can be stored in the local disk, or in a distributed Hadoop file system, or in a distributed database system. 
%such as Teradata, Greenplum, or Aster.
The UDF is responsible for bringing the data to the node where computation is performed.
Once the computation is finished, the UDF can save the output data to local disk or to other platforms with proper formats.

In MPP mode, the UDF supports the MPP data access method.
The MPP data access method enables~(but discourages) data movement between the computing nodes in the cloud.
Data movement between nodes can be expensive and slow.
Therefore, the ideal situation is to have the computation happen in the worker node that has the data.
The master node is responsible for job scheduling and for aggregating the results.
However, data movement and reshuffling are sometimes required.
Therefore, the UDF allows data movement and reshuffling between worker nodes.
In addition, the UDF allows the client to upload data to the cloud and perform computation in the cloud.

In summary, our universal data feeder~(UDF) allows HPSVM to run on a wide range of platforms successfully.

\section{Experiments}

In this section, we test our HPSVM algorithm, and we apply it to a number of applications.
First, we apply HPSVM to some real-world classification problems and compare it with the R package on several public data sets. 
The results demonstrate that our HPSVM implementation yields accuracies similar to or better than those of R implementation, 
but HPSVM runs much faster than the R implementation on large data sets.
Then, we show that HPSVM scales very well on a very big data set as the number of nodes in a distributed environment is increased.
Finally, we simply compare HPSVM with the Spark \cite{spark2015} implementation.
%Finally, we apply HPSVM to a video mining application.

\subsection{Applications of HPSVM and Comparison with LIBSVM Package in R}

We apply HPSVM to some real-world classification problems and compare it with the LIBSVM package 
in R on several public data sets. These experiments were conducted on a non-distributed system that uses Windows 7, 16GB of RAM, and an i7-4770 processor. Our HPSVM implementation yields accuracies similar to those of R implementation, but the run times of HPSVM are several times faster on large data sets. 
In the following paragraphs, we briefly introduce the data sets that we used.

The Mushroom data set is from the UCI Machine Learning Repository \cite{UCI} and consists of 8,124 total observations. 
We partitioned this data set with an 80/20 split, giving us 6,499 training observations and 1,625 test observations. The target is whether a mushroom is edible, `{\em e}', or poisonous, `{\em p}'.

The Adult data set~(also known as the Census Income data set) is also from the UCI Machine Learning Repository \cite{UCI} and is already partitioned into training and testing sets. The training set size is 32,561 observations, and the testing set is 16,281 observations, for a total of 48,842 observations. The target is whether an adult has an income greater than 50,000 dollars.

The Face data set is from the CBCL face database \cite{mitcbcl}. 
This data set is already partitioned into a training set of 2,429 faces and 4,548 non-faces, and a testing set of 472 faces and 23,573 non-faces. The target is whether or not an image is a face.

We present the overall timing and accuracy (correct classification) in Table 1.

\begin{table} [!h]
    \centering
%    \caption{ COMPARISON OF R AND HPSVM }
    \caption{ Comparison of R and HPSVM }
    \begin{tabular} {|m{1.1cm}|m{0.8cm}|m{0.7cm}|m{0.7cm}|m{0.7cm}|m{0.8cm}|m{0.8cm}|}
        \hline
        Data Set Name & Features & Nobs & R Time (sec) & R Accuracy & HPSVM Time (sec) & HPSVM Accuracy \\
        \hline
        Mushroom Train & 22	&6,499	&1.23	&100.0	&{\bf 0.96}	&100.00      \\
        \hline
        Mushroom Test & 22	&1,625	&{\bf 0.05}	&100.0	&0.09	&100.00      \\
        \hline
        Face Train &361 &6,977	&{\bf 6.47}	&{\bf 99.89}	&9.53	&99.36      \\
        \hline
        Face \hspace{10mm} Test  &361 &24,045	&5.37	&97.33	&{\bf 1.11}	&{\bf 97.42}      \\
        \hline
        Adult Train &14 &32,561	&77.30	&85.17	&{\bf 9.88}	&{\bf 85.26}      \\
        \hline
        Adult Test &14 &16,281	&5.65	&85.25	&{\bf 0.31}	&{\bf 85.27}      \\
        \hline
    \end{tabular}
\end{table}

You can see that as the number of observations increases the relative speed of our interior point SVM becomes much faster as compared to the libsvm. It is worthwhile to note that in the Face data set, the libsvm training was faster than our interior point HPSVM implementation. Recalling our CPU time analysis for a single machine, we see that the run time is $O(k_0nm^2/q)$. This scales linearly with $n$, which allows for quick computations as the data size increases in observations. Our implementation scales with the square of the number of features, and thus for this small data example, the libsvm trains faster than our implementation.

%
%
%\subsection{Display Advertising Challenge}
%
%%Tested on the Kaggle Display Advertising Challenge dataset from Criteo
%%(http://labs.criteo.com/2015/03/criteo-releases-its-new-dataset/), which has 4 billion observations and the total data size is of 1TB.
%
%%For our algorithm, the total training time is less than 300 seconds with accuracy value 96.75%.
%
%We evaluated our framework on the Kaggle Display Advertising Dataset (http://labs.criteo.com/2015/03/criteo-releases-its-new-dataset/).
%The data has 4 billions of observations and the data size is about 1TB. By using our implementation, the training can be finished in about 300 seconds and achieves the overall accuracy of 96.75\%. 
%
%

\subsection{Scalability of HPSVM}

In this experiment, we demonstrate that our HPSVM algorithm scales well as the number of computing nodes increases. 
This is very important for training in large data sets.
We apply our HPSVM algorithm to a data set that has approximately 84 million observations and contains 715 features. 
The computation is in a distributed environment, and we show the timing results from changing the number of nodes that we use to run our HPSVM algorithm.
The result is presented in Table 2.

\begin{table} [!h]
    \centering
    \caption{HPSVM Training Times in a Distributed Environment}
    \begin{tabular} {|m{2.33cm}|m{2.33cm}|}
%        \hline
%        \multicolumn{2}{ | c|} {Timing in Seconds for Distributed SVM} \\
        \hline
        Number of Nodes & Training Time~(sec) \\
        \hline
        20 & 631.67 \\
        \hline
        60 & 378 \\
        \hline
        100 & 247.33 \\
        \hline
    \end{tabular}
\end{table}

\subsection{Comparison of HPSVM to Spark}

Spark \cite{spark2015} is a popular open-source machine learning library, 
which includes an implementation of SVM that uses the stochastic gradient descent (SGD) algorithm \cite{shalev2007}. We run a test case to compare it with our HPSVM.

We set up the testing environment with five nodes~(machines). Each node has 32 CPUs and 256GB memory. The testing data is called the ``Glass data set''. Here is the simple description of the data.  

The Glass data set recorded numerous measures in a semiconductor manufacturing
stream. The data came from the engineers who work on developing
the optimal semiconductor production environment.
They developed a sophisticated
system that controlled a large number of variables, such as temperature,
air pressure, air humidity, and so on. In the experiment, the engineers adjusted
the environmental conditions and then checked to see whether the semiconductors
produced under such an environment could satisfy certain requirements.
There are 1001 continuous predictors and 1 million observations in
the Glass data set. The response variable is binary
(whether a semiconductor product passes the test or not), and all the predictors
are standardized to the same scale before training.

We list the overall timing and accuracy (correct classification) in Table 3. From this table, you can see that HPSVM runs faster than Spark and achieves better results.

\begin{table} [!h]
    \centering
    \caption{Comparison of HPSVM to Spark}
    \begin{tabular} {|m{2.00cm}|m{2.33cm}|m{2.33cm}|}
        \hline
        &   Training Time~(sec) & Accuracy~(\%) \\
        \hline
        SPARK & 247 & 90.5\% \\
        \hline
        HPSVM & {\bf 69}  & {\bf 99.78}\% \\
        \hline
    \end{tabular}
\end{table}

\section{Conclusion}
In this paper, we present a high-performance support vector machines~(HPSVM) algorithm that scales well in large data sets. We implemented this algorithm with MPI. The implemented algorithm is now running on various systems, including a distributed Hadoop file system and a distributed database system~(such as Teradata, Greenplum, Aster, and so on). We compare the accuracy of our implementation with the state-of-the-art SVM technique implemented in R on some public data sets. When the data set is large, experiments show that our algorithm scales very well and generates better models.
%In the further, we will investigate kernel based algorithm with distributed compuation and scoring.

%%%%%%%%%%%%%%%%%%%%%%%%%%%%%%%%%%%%%%%%%%%%%%%%%%%%%%%%%%%%%%%%%%
%%%%%%%%
%%%%%%%% Reference
%%%%%%%% Below is an example of bibliography that contains all entries within this document.
%%%%%%%% You can also let BibTeX generate your bibliography by inserting the following two commands:
%%%%%%%%
%%%%%%%% \bibliographystyle{IEEEtran}
%%%%%%%% \bibliography{<your_bibliography_file_1>,<your_bibliography_file_2>,...}
%%%%%%%%
%%%%%%%% Note that you need to make sure that LaTeX (BibTeX) can find IEEEtrans.bst in your system.
%%%%%%%% If you are unsure about that, just place IEEEtrans.bst in the same directory where your LaTeX source files reside.
%%%%%%%%
%%%%%%%%%%%%%%%%%%%%%%%%%%%%%%%%%%%%%%%%%%%%%%%%%%%%%%%%%%%%%%%%%%%
%%%%%%%%% Below thebibliography environment will be automatically created in a different file (your_file_name.bbl) 
%%%%%%%%% if you use BibTeX and specify IEEEtrans.bst.
%%%%%%

\end{document}